\documentclass[letterpaper, 10 pt, conference]{ieeeconf}  

\IEEEoverridecommandlockouts                              

\usepackage[pdftex]{graphicx}
\usepackage[percent]{overpic}
\usepackage{multirow}
\usepackage{verbatim}
\usepackage{colortbl}
\usepackage{wrapfig}
\usepackage{soul}
\usepackage{url}
\usepackage{algorithm}

\usepackage[noend]{algpseudocode}
\usepackage{amsmath}
\usepackage{amssymb}
\usepackage{mathtools}
\usepackage{overpic}
\usepackage{longtable}
\usepackage{float}
\usepackage{multicol}
\usepackage{hyperref}
\usepackage{textcomp}

\usepackage{pdfpages}

\usepackage[normalem]{ulem}

\usepackage[sc]{mathpazo}

\usepackage{xcolor}

\newtheorem{thm}{Theorem}[section]

\newtheorem{remark}[thm]{Remark}

\newenvironment{myitem}{\begin{list}{$\bullet$}
{\setlength{\itemsep}{-0pt}
\setlength{\topsep}{0pt}
\setlength{\leftmargin}{12pt}
\setlength{\parsep}{0pt}
\setlength{\itemsep}{0pt}
\setlength{\partopsep}{0pt}}}%
{\end{list}}

\newif\ifremark
\long\def\remark#1{
\ifremark%
   \begingroup%
   \dimen0=\textwidth
   \advance\dimen0 by -1in%
   \setbox0=\hbox{\parbox[b]{\dimen0}{\protect\em #1}}
   \dimen1=\ht0\advance\dimen1 by 2pt%
   \dimen2=\dp0\advance\dimen2 by 2pt%
   \vskip 0.25pt%
   \hbox to \textwidth{%
      \vrule height\dimen1 width 3pt depth\dimen2%
      \hss\copy0\hss%
      \vrule height\dimen1 width 3pt depth\dimen2%
   }%
   \endgroup%
\fi}

\definecolor{Red}{rgb}{0.5,0.1,0.1}

\DeclarePairedDelimiterX{\norm}[1]{\lVert}{\rVert}{#1}
\newcommand{\rom}[1]{\uppercase\expandafter{\romannumeral #1\relax}}

\definecolor{myblue}{rgb}{0.15, 0.05, 0.7}

\newtheorem{blemma}[thm]{\bf Lemma}
\newtheorem{coro}[thm]{\bf Corollary}

\usepackage{xargs}

\title{\LARGE \bf
Resolution Complete In-Place Object Retrieval\\given Known Object Models
}

\author{Daniel Nakhimovich, Yinglong Miao, and Kostas E. Bekris
\thanks{ The authors are with the Dept. of Computer Science, Rutgers, New Brunswick, NJ. Email: {\tt \{d.nak, ym420,kb572\}@rutgers.edu}. The work is partially supported by NSF awards 1934924 and 2021628. The opinions expressed here are those of the authors and do not reflect the positions of the sponsor.}%
}

\begin{document}

\maketitle
\thispagestyle{empty}
\pagestyle{empty}

\begin{abstract}
This work proposes a robot task planning framework for retrieving a target object in a confined workspace among multiple stacked objects that obstruct the target. The robot can use prehensile picking and in-workspace placing actions.
The method assumes access to 3D models for the visible objects in the scene. The key contribution is in achieving desirable properties, i.e., to provide (a) safety, by avoiding collisions with sensed obstacles, objects, and occluded regions, and (b) resolution completeness ({$\tt RC$}) - or probabilistic completeness ({$\tt PC$}) depending on implementation - which indicates a solution will be eventually found (if it exists) as the resolution of algorithmic parameters increases. A heuristic variant of the basic {$\tt RC$} algorithm is also proposed to solve the task more efficiently while retaining the desirable properties. Simulation results compare using random picking and placing operations against the basic {$\tt RC$} algorithm that reasons about object dependency as well as its heuristic variant. The success rate is higher for the {$\tt RC$} approaches given the same amount of time. The heuristic variant is able to solve the problem even more efficiently than the basic approach. The integration of the $\tt RC$ algorithm with perception, where an RGB-D sensor detects the objects as they are being moved, enables real robot demonstrations of safely retrieving target objects from a cluttered shelf.
\end{abstract}

\section{INTRODUCTION}

Robotic manipulation has the potential of being integrated into daily lives of people, such as in household service areas \cite{wong2013manipulation, garrett2020online}. A useful skill for such household settings involves the retrieval of a target object from a confined and cluttered workspace, such as a fridge or a shelf, which may also require the rearrangement of other objects in the process. In this context, it is important to consider how to safely retrieve objects while minimizing the time spent or the amount of pick and place operations, so as to assist humans efficiently. 

\begin{figure}[thpb]
  \centering
  \includegraphics[width=\linewidth,
  trim={0 3cm 0 1cm},clip]{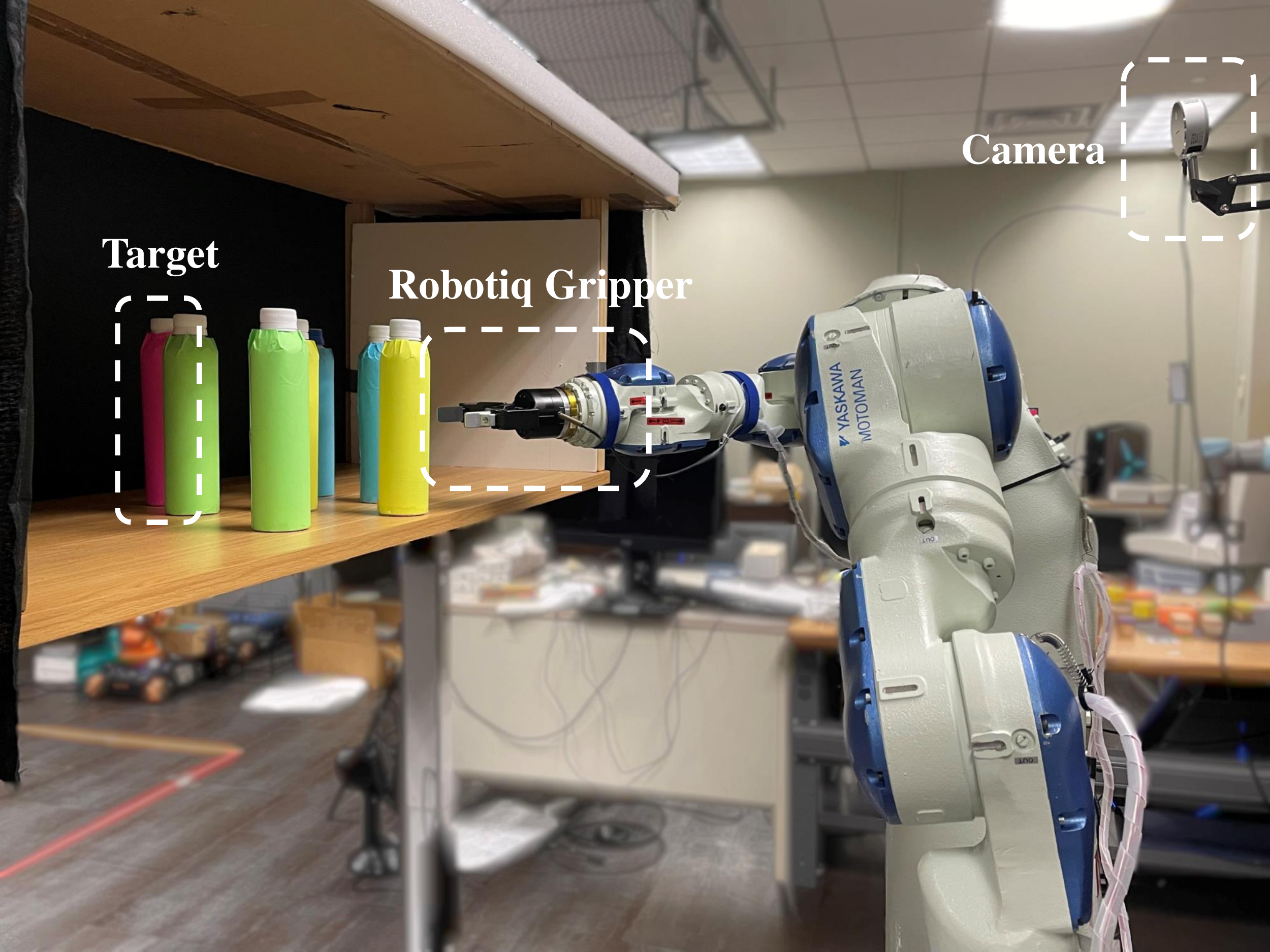}
  \includegraphics[width=0.465\linewidth,trim={13.9cm 0 12cm 12.9cm},clip]{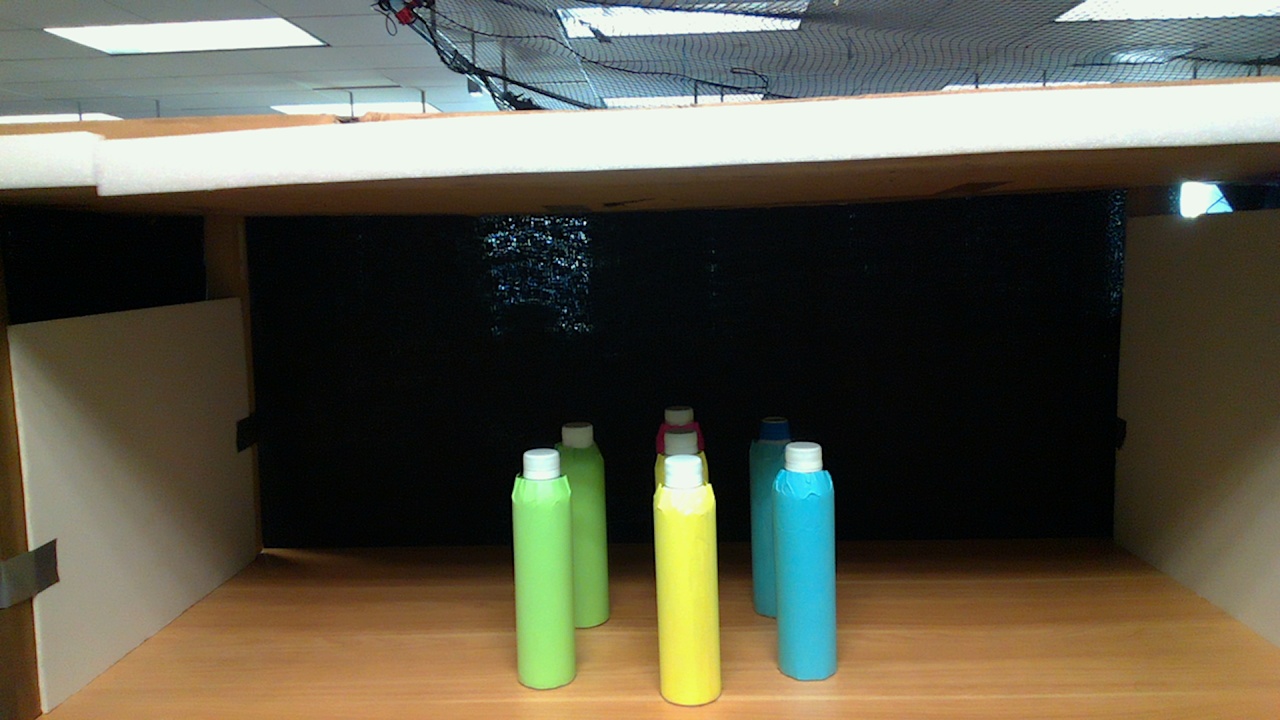}
  \includegraphics[width=0.52\linewidth,trim={4cm 7cm 5cm 2cm},clip]{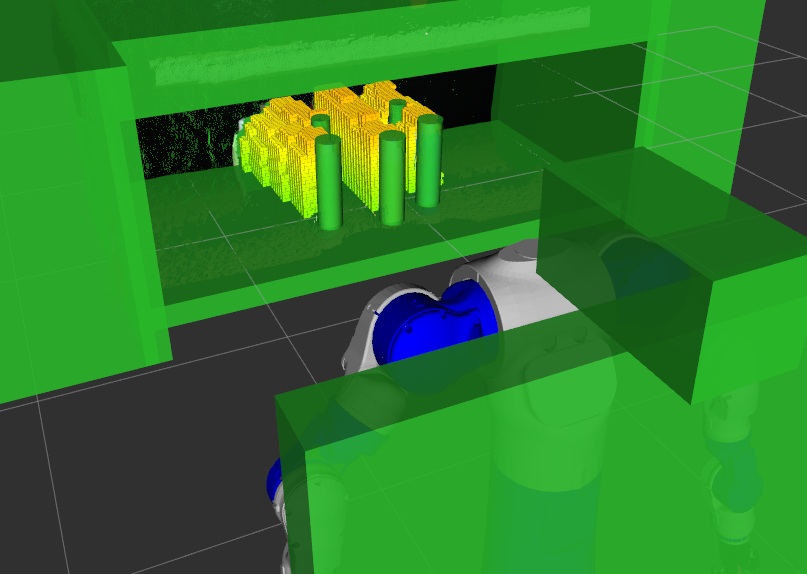}
  \vspace{-0.3in}
  \caption{(Top) Setup for the real demonstration using an RGB-D sensor, robotiq gripper, and Yaskawa Motoman robot to retrieve the target bottle. (Bottom Left) The camera view in which objects are occluded. (Bottom Right) The corresponding voxel map.}
  \label{fig:real_setup}
  \vspace{-0.32in}
\end{figure}

One of the challenging aspects of these problems that requires explicit reasoning relates to heavy occlusions in the scene, as the sensor is often mounted on the robot and has limited visibility. These visibility constraints complicate the task planning process, as rearranging one object can limit placements for others and can introduce new occlusions. Moreover, real-world scenes in household setups are often unstructured and involve objects with complex spatial relationships, such as objects stacked on each other.

Many previous efforts on object retrieval have focused on cases where blocking objects are extracted from the workspace \cite{dogar2014object, nam2021fast}, which simplifies the challenge as it does not require identifying temporary placement locations for the objects within the confined space. In-place rearrangement has been considered in some prior efforts \cite{ahn2021integrated}. While this prior method is efficient, it is not complete as it limits the reasoning on the largest object in the scene to analyze object traversability \cite{nam2021fast}. Alternatives use machine learning to guide the decision making \cite{bejjani2021occlusion, huang2022mechanical, price2019inferring}, which is an exciting direction but does not easily allow for performance guarantees, such as resolution completeness. Setups where object stacking arise have received less attention and most solutions that do consider stacking are dependent on machine learning for reasoning \cite{huang2022mechanicalstack, zhu2021hierarchical, kumar2022graph}. Some works have proposed testbeds \cite{liu2021ocrtoc} that can help evaluate solutions in this domain.

This work focuses on object retrieval in clutter where occlusions arise and objects may be  initially stacked under the assumption of known object models (e.g. \autoref{fig:dep_graph}). It aims at a theoretical understanding to show the algorithm has safety and $\tt RC$ guarantees. A heuristic variant improves the practical efficiency. Key features of the proposed $\tt RC$ framework are the following:


\begin{myitem}
    \item it employs an adaptive dependency graph data structure inspired by solutions in object rearrangement with performance guarantees \cite{wang_uniform_2021} that express a larger variety of object relationships than previously considered (namely occlusion dependencies);
    \item it computes the occlusion volume of detected objects as a heuristic to inform the planning process;
    \item it reasons about the collision-free placement of objects in the confined workspace efficiently by utilizing a voxelized representation of the space;
    \item it achieves $\tt RC$ (or $\tt PC$) depending on the implementation of the underlying sampling subroutines;
    \item it provides an early termination criterion when a solution cannot be found for the given resolution. 
\end{myitem}

Simulation results, using a model of a Yaskawa Motoman manipulator for rearranging objects on a tabletop as shown in Fig. \ref{fig:real_setup}, evaluate the proposed {$\tt RC$} framework against a baseline using random picking and placing operations. Both variants of the  {$\tt RC$} reason about object dependencies. One doesn't use heuristics and one is heuristically guided while retaining {$\tt RC$}. Both of the {$\tt RC$} approaches outperform the baseline. The heuristically guided solution is able to solve the problem more efficiently than the basic {$\tt RC$} solution. The integration of the proposed approach with perception, where an RGB-D sensor detects the objects as they are being moved, provides real robot demonstrations of safe object retrieval from a cluttered shelf.

\section{RELATED WORK}




Some works on object retrieval rely on geometric analysis of object occlusion \cite{dogar2014object,nam2021fast}. They provide theoretical insights but frequently do not limit actions to in-place rearrangement of blocking objects. Specifically, one method constructs a dependency graph taking into account objects that jointly occlude a region and objects that block others \cite{dogar2014object}. The occlusion volume is used to estimate belief regarding the target object position and helps to construct an optimal A* algorithm. An alternative constructs a Traversability graph (T-graph) \cite{nam2021fast}, where the edges encode if the largest object in the scene can be moved between two poses. It then constructs an  algorithm to extract the target object, but is limited as the traversability edges are too constraining. The POMDP formulation is popular for the task \cite{zhao2021hierarchical, xiao2019online}, which allows the application of general POMDP solvers. The POMDP formulation was also adopted by the work that formalizes object retrieval in unstructured scenes as "mechanical search" \cite{danielczuk2019mechanical}.

Alternatives rely on learning-based methods to solve such challenges, such as reinforcement learning \cite{bejjani2021occlusion} or target belief prediction \cite{huang2022mechanical, huang2022mechanicalstack, huang2021mechanical}. They report good performance but do not provide theoretical guarantees given the black-box nature of the solutions. In particular, a reinforcement learning solution \cite{bejjani2021occlusion} uses the rendered top-down projections of the scene to predict the target poses. A recent follow-up effort \cite{huang2022mechanical} on previous work \cite{huang2021mechanical} estimates the 1D position belief of the target object on the shelf via machine learning. It then constructs a policy based on the distribution change after applying pushing and suction actions. It incorporates stacking and unstacking actions, where object stacking is represented by a tree structure. Other works such as \cite{zeng2022robotic} utilize learning for planning grasps to greedily empty bins of complex and novel objects.

A related work \cite{miao2022safe} proposes a complete framework to safely reconstruct all objects in the scene amidst object occlusions. Nevertheless, object retrieval may not require reconstructing all objects and requires a search procedure that is more task-driven for efficiency. There are also previous works \cite{gupta2013interactive, miao2022safe} that construct a voxelization of the environment to model object occlusions, similar to the current work. This representation is used to compute an object's occlusion volume, which provides heuristic guidance. Object spatial relationships are often represented by scene graphs \cite{zhu2021hierarchical, kumar2022graph}, or implicitly in machine learning solutions \cite{danielczuk2019mechanical, poon2019probabilistic, novkovic2020object}.

What stands out in this work is that it proposes a general template for a $\tt RC$ or $\tt PC$ approach to task retrieval in occluded environments that only relies on basic motion and perception primitives. This modular nature allows for quick sim-to-real transfer and passive performance improvement as the primitives are improved over time.
Furthermore, an efficient implementation is demonstrated utilizing a
voxelized representation of the environment for quick collision filtering of object placements as well as providing an effective heuristic to rank object manipulations. Thus, the framework enables effective in-workspace manipulation.

\section{PROBLEM STATEMENT}

\newcommand{\nil}{{\tt null }}
\newcommand{\seE}{\mathbf{SE(3)}}
\newcommand{\obdb}{\mathbb{O}} 
\newcommand{\obe}{o} 
\newcommand{\obc}{s} 
\newcommand{\obt}{d} 
\newcommand{\os}{O} 
\newcommand{\dos}{\tilde{O}} 
\newcommand{\obis}{\mathbb{S}} 
\newcommand{\obd}{\mathbb{D}} 
\newcommand{\mplan}[1]{{\tt MotionPlanner(#1)}}
\newcommand{\pick}[1]{{\tt MPPick(#1)}}
\newcommand{\place}[1]{{\tt MPPlace(#1)}}


Consider an environment with a set $\obis = \{\obe_1, ..., \obe_n\} \subset \obdb$ of $n$ objects for which there are available 3D models. The objects are stably resting on a support surface. Objects are allowed to be initially stacked and occlude each other from the camera view.

The robot has one fixed RGB-D sensor at its disposal. 
Discovered objects are those recognized given the observation history. An objects is assumed to be recognized once an image segmentation process recognizes it as an individual object in the observed image. 
Similarly, a perception method for detecting the target object once observed is assumed. The region of the workspace occluded by object $\obe_i$ at pose $\obc_i$ is denoted as $\os_i(\obc_i)$. Similarly the space uniquely occluded by object $\obe_i$ at pose $\obc_i$, called the direct occlusion space, is denoted as $\dos_i(\obc_i)$. 
The proposed algorithms gradually removes occlusions and recognizes objects. A motion planner is used to plan pick-and-place actions.




While objects can start out stacked, they are not re-stacked and are only placed on the ground surface during actions.
While objects can start out stacked, no reasoning about stability and ability to re-stack objects is considered. Thus, once an action to pick up a stacked object is taken, that object will only be placed on the ground surface.
For further assumptions and required properties of the motion planner see \autoref{sec:completeness}.

The objective for the object retrieval task is to determine a sequence of pick and place actions in order to discover and subsequently retrieve the target object; the target need not be directly visible or pickable from the robot's sensors. The corresponding solution should provide desirable guarantees: (a) safety, by avoiding collisions with sensed obstacles and objects as well as occluded regions, and (b) resolution completeness ({\tt  RC}) - or alternatively probabilistic completeness, depending on the implementation of the underlying motion planner, grasping process and object placement sampling.
The optimization objective is to minimize the number of performed actions until the target object is retrieved.


\section{METHOD}

\newcommand{\depgraph}[1]{{\tt DepGraph(#1)}}
\newcommand{\placement}[1]{{\tt SamplePlacement(#1)}}
\newcommand{\addHiddenEdges}[1]{{\tt AddHiddenEdges(#1)}}


The proposed pipeline is detailed in
\autoref{alg:pipeline}. First a voxelized representation of the scene and a dependency graph are computed (lines 3,4,5), which are detailed in \autoref{subsec:voxel_dep}. The dependency graph contains a belief state of the current scene based on visibility and reachability constraints. All the \textit{sinks} of this directed graph represent likely pickable objects.
The \textit{ranks} are described later (see \autoref{subsec:target-prediction}).
If the target object is pickable then it is retrieved and the pipeline is terminated line 6.
Otherwise, a placement is planned (see \autoref{subsec:placement_sampling}) one at a time for each object in a shuffled order biased by the \textit{ranks} (\textit{TryMoveOne} line 7). The first successful plan is executed.

If no placement is found for any of the pickable objects, then a fallback procedure is called
to try and pick one object and move it temporarily out of view of the camera; the first successful plan is executed, the scene is re-sensed, and a new placement is sampled for the object or the object is simply put back at the same spot (\textit{MoveOrPlaceback} line 10). At this stage, the pipeline could restart if a new object was discovered (lines 12-15). Otherwise the same set of pickable objects 
are tested for new placements in the scene (line 16). If these two operations of ``moving an object to look behind it''\textit{(MoveOrPlaceback)} followed by retrying to ``move one of the pickable objects to a new spot''\textit{(TryMoveOne)} fail for all objects, then the pipeline can return and report failure for the current resolution (line 22). If the sequence of operations succeeds, then the pipeline can restart (line $19\rightarrow2$).

\newcommand{\trymoveone}[1]{{\tt TryMoveOne(#1)}}
\newcommand{\move}[1]{{\tt MoveOrPlaceback(#1)}}
%
%
%

\begin{algorithm}[thpb]
\caption{$\tt RC\_Pipeline$($\tt target$)}\label{alg:pipeline}
\begin{algorithmic}[1]
    \State $\tt failure \gets false$
    \While{$\tt failure = false$}
        \State $\tt space \gets UpdateVoxelsFromImage()$
        \State $\tt dg \gets \depgraph{space}$
        \State $\tt sinks,ranks \gets RankSinks(target,dg)$ 
        \If{$\tt target \in sinks$} $\tt break$\EndIf
        \If{$\tt \trymoveone{sinks,ranks} = false$}
            \State $\tt failure \gets true$
            \For{$\tt sink \in sinks$}
                \If{$\tt \move{sink} = false$}
                    \State $\tt continue$
                \EndIf
                \State $\tt space \gets UpdateVoxelsFromImage()$
                \If{$\tt DidDiscoverObject(space)$}
                    \State $\tt failure \gets false$
                    \State $\tt break$
                \EndIf
                \If{$\tt \trymoveone{sinks,\emptyset} = false$}
                    \State $\tt continue$
                \EndIf
                \State $\tt failure \gets false$
                \State $\tt break$
            \EndFor
        \EndIf
    \EndWhile
    \If{$\tt not~failure$} 
        \State $\tt Retrieve(target)$
    \EndIf
    \State $\tt return~failure$
    
\end{algorithmic}
\end{algorithm}

\begin{figure*}[thpb]

    \begin{overpic}[width=0.336\linewidth,trim={0 140 40 22},clip]
    {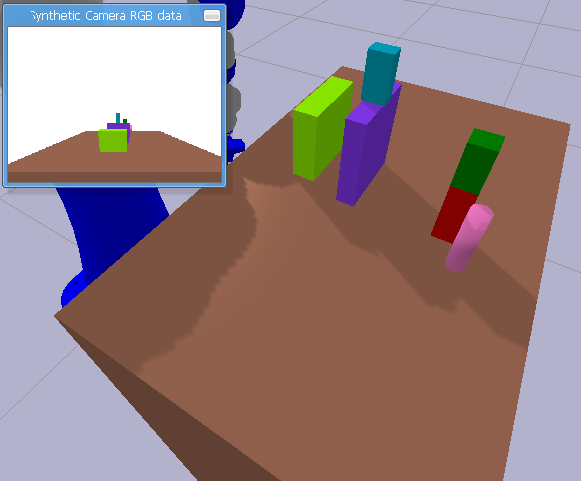}
        \put (3,10) {\huge$\displaystyle 1)$}
    \end{overpic}
    \begin{overpic}[width=0.336\linewidth,trim={0 140 40 22},clip]
    {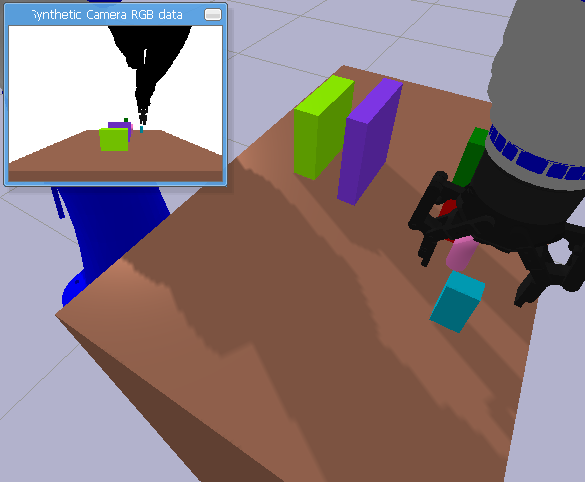}
        \put (3,10) {\huge$\displaystyle 2)$}
    \end{overpic}
    \begin{overpic}[width=0.336\linewidth,trim={0 140 40 22},clip]
    {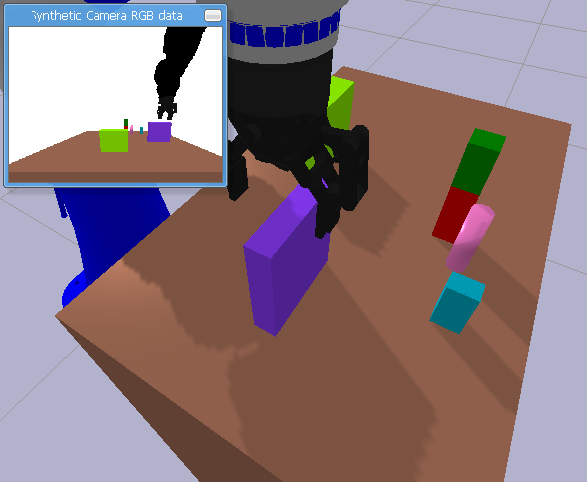}
        \put (3,10) {\huge$\displaystyle 3)$}
    \end{overpic}
    \begin{overpic}[width=0.336\linewidth,trim={0 140 40 22},clip]
    {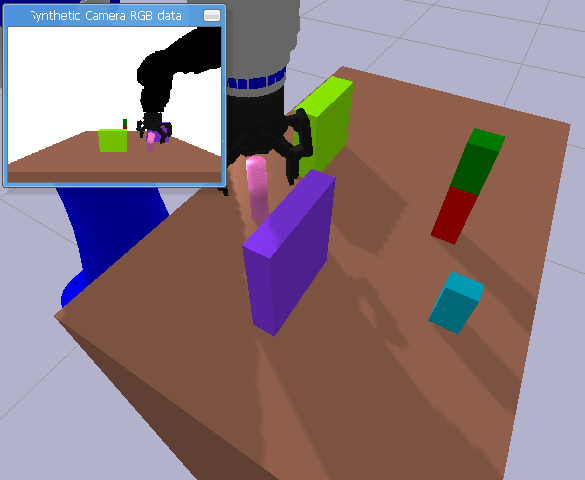}
        \put (3,10) {\huge$\displaystyle 4)$}
    \end{overpic}
    \begin{overpic}[width=0.336\linewidth,trim={0 140 40 22},clip]
    {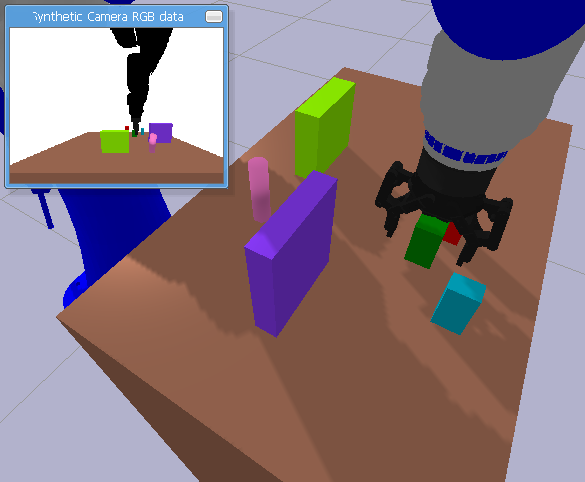}
        \put (3,10) {\huge$\displaystyle 5)$}
    \end{overpic}
    \begin{overpic}[width=0.336\linewidth,trim={0 140 40 22},clip]
    {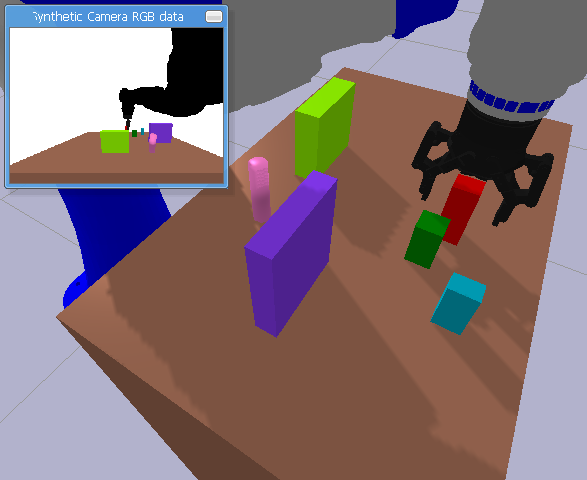}
        \put (3,10) {\huge$\displaystyle 6)$}
    \end{overpic}
    \begin{overpic}[width=0.245\linewidth]
    {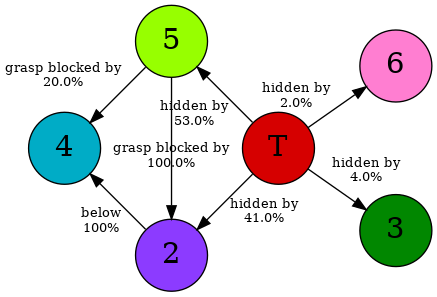}
        \put (-1,5) {\huge a)}
    \end{overpic}
    \begin{overpic}[width=0.245\linewidth]
    {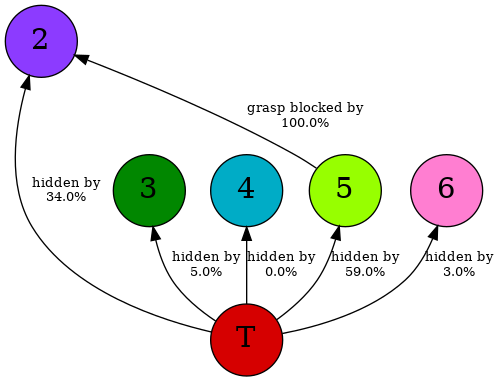}
        \put (-1,5) {\huge b)}
    \end{overpic}
    \begin{overpic}[width=0.245\linewidth]
    {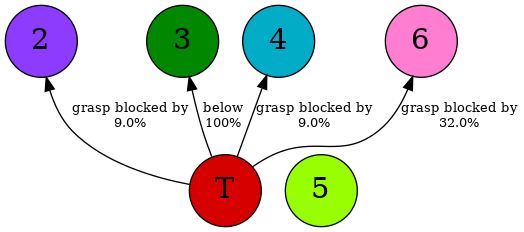}
        \put (-1,5) {\huge c)}
    \end{overpic}
    \begin{overpic}[width=0.245\linewidth]
    {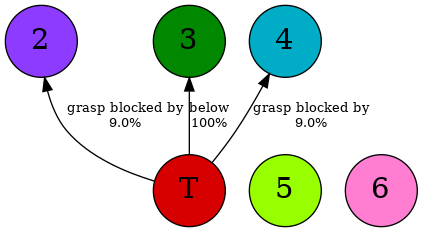}
        \put (-1,5) {\huge d)}
    \end{overpic}
  \caption{
  Images (1)-(6) show a simulated experiment from initial configuration to final one action at a time with corresponding camera views in the top left. The corresponding generated dependency graphs transitioning between the images (1)-(2), (2)-(3), (3)-(4), and (4)-(5) are shown in images (a)-(d). 
  The colors of nodes corresponds with the objects in the scene and the red object (labeled `T') is the target object for the trial.
  The last graph between images 5-6 is not shown since it is trivial having no dependencies between any objects.
  }
  \label{fig:dep_graph}
  \vspace{-0.28in}
\end{figure*}

\subsection{Voxel Map and Dependency Graph}
\label{subsec:voxel_dep}

A 3D occlusion voxel grid within the workspace is constructed from RGB-D images.
First the point cloud (in world frame) is generated using the RGB-D image and the inverse of the camera projection. These points are down-sampled into a voxelization of the scene.
Given segmentation image information, each object is associated with a portion of the voxel grid. Object geometry is used to label voxels as occupied and remaining associated voxels as occluded.
The occluded regions of objects may intersect when jointly occluded.

The dependency graph is a directed graph where each node represents a visible (or target) object and a labeled edge $(\obe_i, \obe_j, r)$ represents a relation $r$ from object $\obe_i$ to $\obe_j$ that necessitates $\obe_j$ to be picked and placed before $\obe_i$ could be picked.
Valid relations in this work include ``below'', ``grasp blocked by'', and -- for the prediction of the target object -- ``hidden by''. See \autoref{fig:dep_graph} for a sequence of such dependency graphs generated during an example experiment. 

Object x is defined to be ``below'' object y ($x \xrightarrow{below} y$) if object x touches object y and the z-coordinate of the center of mass of x is less than that of y. Note that this isn't guaranteed to capture all intuitive cases of one object being below another for non-convex objects. This relation is computed using object models and poses given by the perception system.

Object x has its ``grasp blocked by'' y ($x \xrightarrow{blocked} y$) if there are no collision free grasp poses for object x and the arm is in collision with y for one or more valid grasp poses.
(Grasp poses are sampled and tested by inverse kinematics (IK) for discovered objects)
Grasp poses are sampled using inverse kinematics (IK) to discovered objects. 
(or, if the grasping pose collides with objects, an edge to each collided object is added)
If there exists a collision free grasp for an object, no blocking edges are added; otherwise, an edge to each object that has a collision with the arm is added. Note that although this relation guarantees that the source object blocks the target, it doesn't capture all such reachability dependencies. This however is not an issue for completeness as \autoref{alg:pipeline} will eventually try to grasp all objects in the case of motion planning failure.

The target object t is possibly ``hidden by'' x ($t \xrightarrow{hidden} x$) if the target isn't sensed in the scene and object x is touching the table. This relation is used to keep track of the belief state of where the target is. Each edge is assigned a probability based on the volume of the occluded space behind object x (see \autoref{subsec:target-prediction}).



\vspace{-0.11in}
\subsection{Placement Sampling}
\label{subsec:placement_sampling}
A valid placement is one that doesn't collide with another object or any undiscovered area of the workspace.
Instead of randomly sampling x,y coordinates for an object and checking for collision at that point, we create a grid matching the horizontal extents of the workspace and add a collision mask which is the shadow of the object occupancy and occlusion voxel grid looking from a birds eye view. This mask is then convolved with the shadow of the object that is to be placed. 
The occupied pixel indices indicate collision-free placements and are converted to world coordinates.
Object orientations can be enumerated by rotating the object shadow.



\subsection{Target Object Prediction}
\label{subsec:target-prediction}
Intelligent object location prediction is achieved by applying a heuristic which ranks the pickable objects determined by the dependency graph (line 4 in \autoref{alg:pipeline}).
This is done by augmenting the dependency graph edges with a weight $p \in (0,1]$ estimating the probability for the relation $\obe_i \overset{r}{\longrightarrow} \obe_j$ to be true.
To rank a pickable object, the sum of products of edge weights of all simple paths between the target and the object is computed. When sampling from the list of pickable objects, this rank is used as a probability weight.

For the ``below'' relation $p=1$ since object segmentation is assumed to be reliable. For the ``grasp blocked by'' relation p is equal to the fraction of total sampled grasps for which that object is in collision with the arm. Note, the weights of grasp blocking edges coming out of any object need not add to one (hence don't truly represent probabilities) since the arm could be colliding with multiple objects for any single grasp.
For the ``hidden by'' relation, the goal is to encourage knowledge gain of the environment. This is done by normalizing the volume of the direct occlusion region of each stack of objects and assigning the inverse as the probability estimate that the target is hidden behind each stack. This heuristic biases the pipeline towards discovering large volumes of occluded workspace.

From an algorithmic point of view, there is technically no reason to normalize the output of the heuristic, however, representing the heuristics as probabilities is insightful since - without prior knowledge - the probability the hidden object to be in a larger volume is larger than the probability of it being in a smaller volume. Furthermore, modeling the dependency graph edges with probabilities as opposed to non-normalized weights is conducive for exploring future work which might seek to combine the probabilities based on the proposed volumetric-heuristics with priors based on the semantics of the objects involved or from an additional human instruction (see \autoref{sec:conclusion} work).

\section{RESOLUTION COMPLETENESS}

\label{sec:completeness}
\newcommand{\algrand}{{\tt RAND-ACT}}
\newcommand{\algdep}{{\tt DEP-ACT}}


Given a (formally) complete motion planner (finds solution in finite time if exists),
a continuous space grasp sampler, and continuous placement sampler, the proposed pipeline would be $\tt PC$
Given a $\tt RC$ motion planner, 
a discrete space grasp sampler, and discrete placement sampler, the proposed pipeline would be $\tt RC$.

To show the $\tt PC$ or $\tt RC$ of the algorithm proposed in this work, a simpler version of the algorithm is analysed first. Without loss of generality, the actual proposed algorithm will be likewise proven complete.

Consider a much simpler algorithm that, at every iteration, tries to pick an object at random and, if is not the target object, subsequently place it randomly in the explored region of the workspace; call this \algrand.

\begin{blemma}
\algrand~is $\tt PC$ or $\tt RC$ (depending on the planning and sampling subroutines).
\end{blemma}
\begin{proof}
At every iteration, \algrand~attempts to perform a random action. Consequently, \algrand~executes a random walk on the space of all actions. Since pick and place actions are reversible, if a sequence of such valid actions exists, this algorithm will eventually perform it (or an augmented version of the sequence - i.e. placing an object back to where it was picked from) in the limit (or finite time for {\tt RC} subroutines).
\end{proof}

\begin{coro}
\autoref{alg:pipeline} is $\tt PC$ or $\tt RC$ (depending on the planning and sampling subroutines)).
\end{coro}
\begin{proof}
Indeed \autoref{alg:pipeline} is really a fancy implementation of \algrand. At every iteration, the dependency graph is used to identify (and heuristically rank) the currently pickable objects. One of these is chosen randomly for a pick and place action via the $\tt TryMoveOne$ subroutine. 
The $\tt MoveOrPlaceback$ subroutine acts as a fallback in case there are no new discovered placements found; it delays placement sampling till after the environment is re-sensed with the picked object moved out of the way.
Notice further, that even if the pickable objects are sampled weighted according to their ranking rather than uniformly, the action space is still explored entirely because each action has a positive probability of being sampled.
Thus, {\bf w.l.o.g. \autoref{alg:pipeline}} is $\tt PC$ or $\tt RC$ as well.
\end{proof} 

{\bf Failure Detection:} 
The implementation in this work uses the $\tt RC$ approach.
In addition to $\tt RC$, \autoref{alg:pipeline} takes a step closer towards achieving general completeness by actually detecting certain unsolvable cases within the completeness constraints of the motion and sampling subroutines.
The detectable unsolvable instances for which the algorithm will return failure in finite time are as follows.
\begin{myitem}
    \item No object can be grasped. This could happen if two objects each block the grasp of the other.
    \item No objects can be placed anywhere (except its current spot). This could happen in a highly cluttered scene where the only valid placement for each object is to just put it back where it was.
\end{myitem}
Thus, \autoref{alg:pipeline} has stronger guarantees than $\tt RC$ but is not formally complete since it may run forever by juggling two objects between two placements.

{\bf A Caveat:} A fundamental assumption in the argument presented is that actions are reversible. This is not always true in practice depending on the implementation of the sampling subroutines. And indeed, the implementation of the placement sampling process proposed in \autoref{subsec:placement_sampling} implies irreversible actions for scenes with stacked objects because it does not consider the possibility of re-stacking objects. Thus, the proposed pipeline (as implemented) is only resolution complete on the any sub-task where the objects are no longer stacked. Implementation of stacking actions is planned for future work.




\section{EXPERIMENTS}

\begin{figure*}[thpb]
\vspace{0.1in}
\includegraphics[width=0.162\linewidth,trim={0 0 0 0},clip]{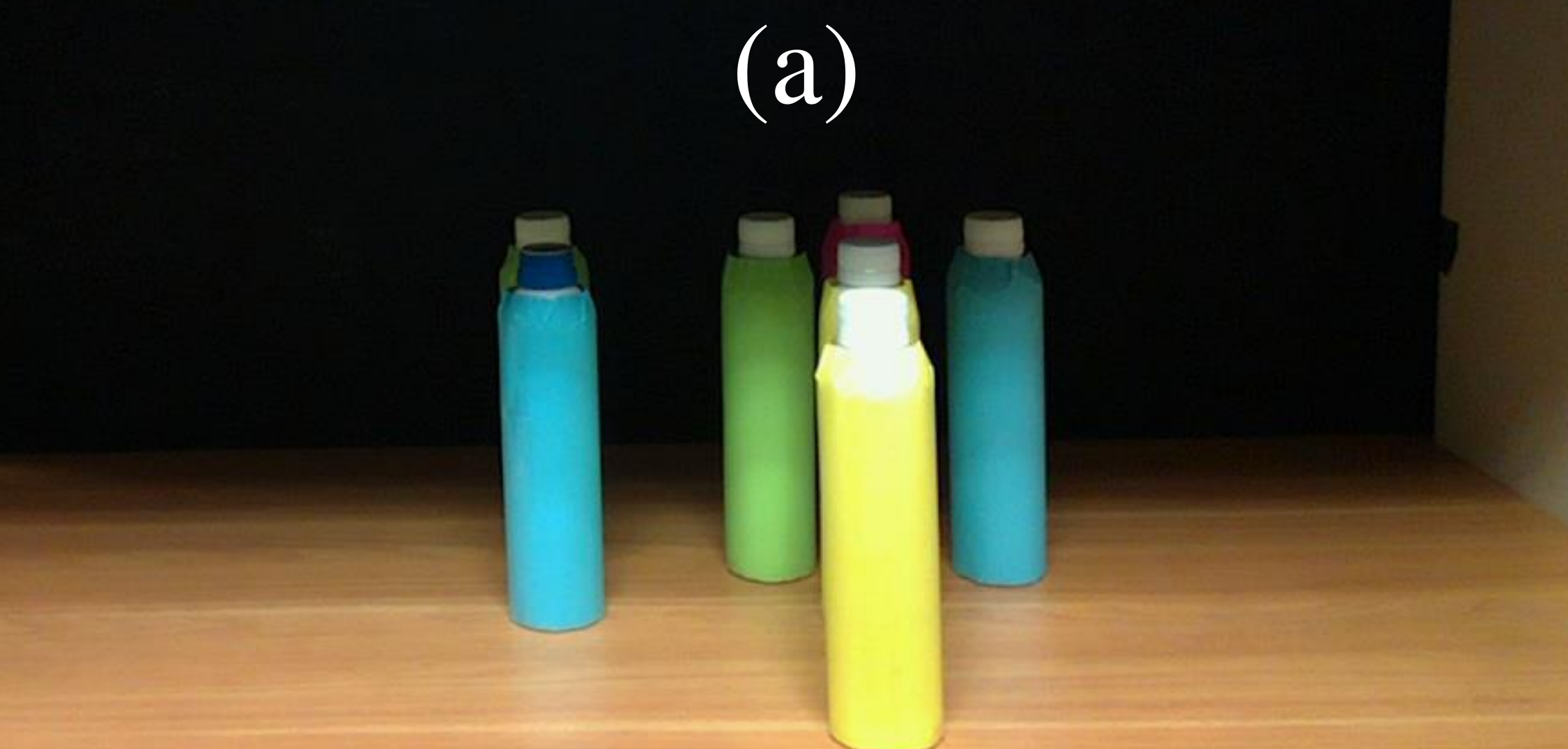}
\includegraphics[width=0.162\linewidth,trim={0 0 0 0},clip]{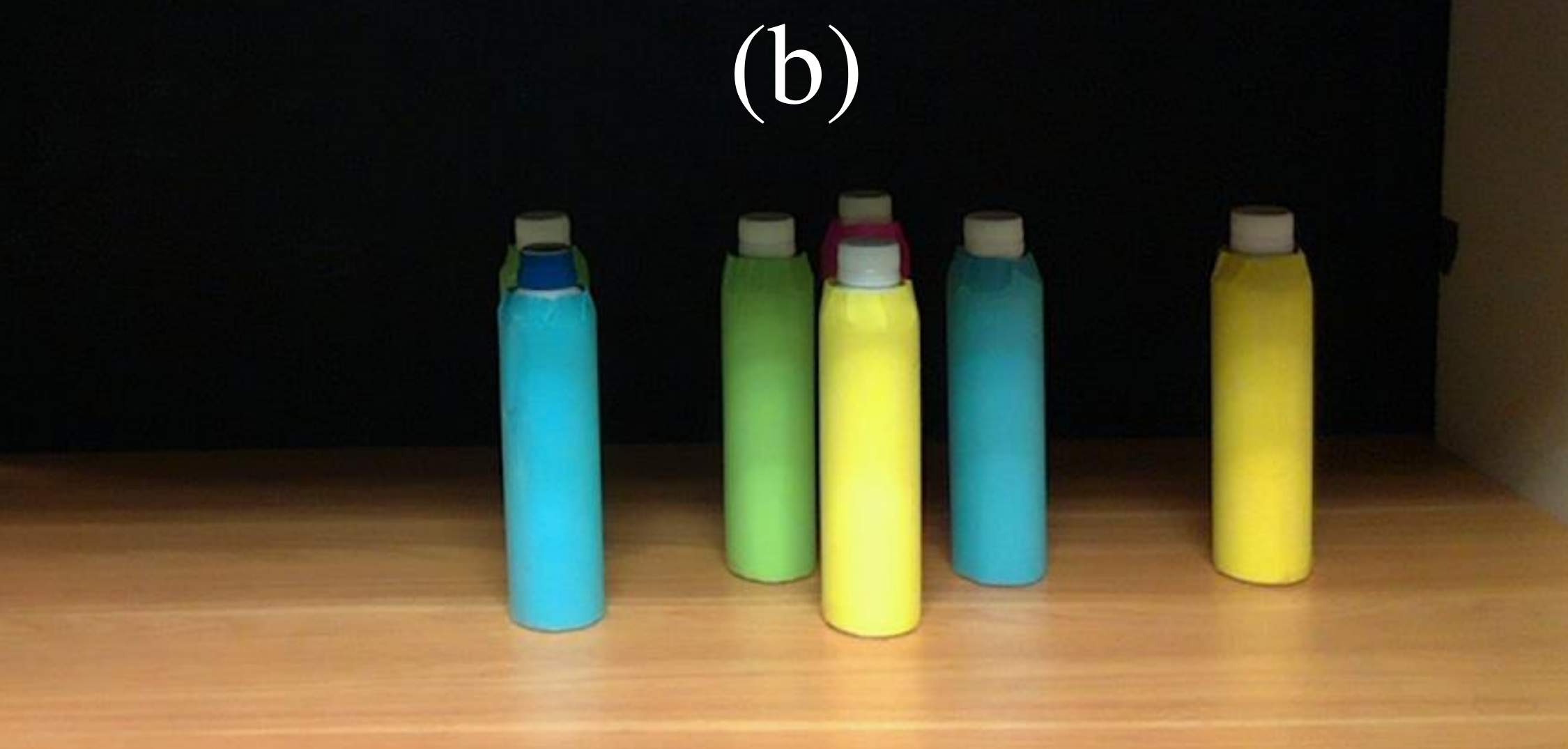}
\includegraphics[width=0.162\linewidth,trim={0 0 0 0},clip]{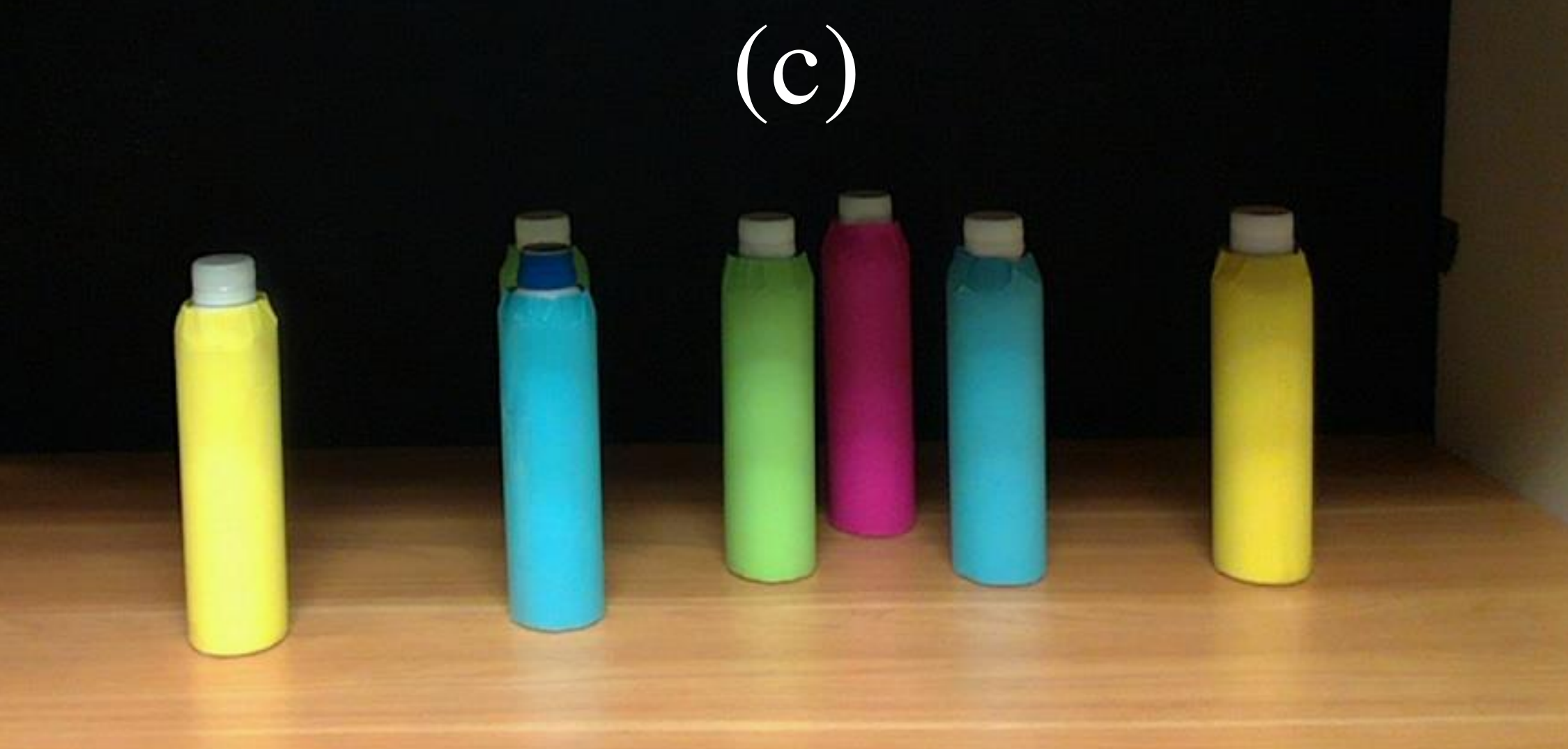}
\includegraphics[width=0.162\linewidth,trim={0 0 0 0},clip]{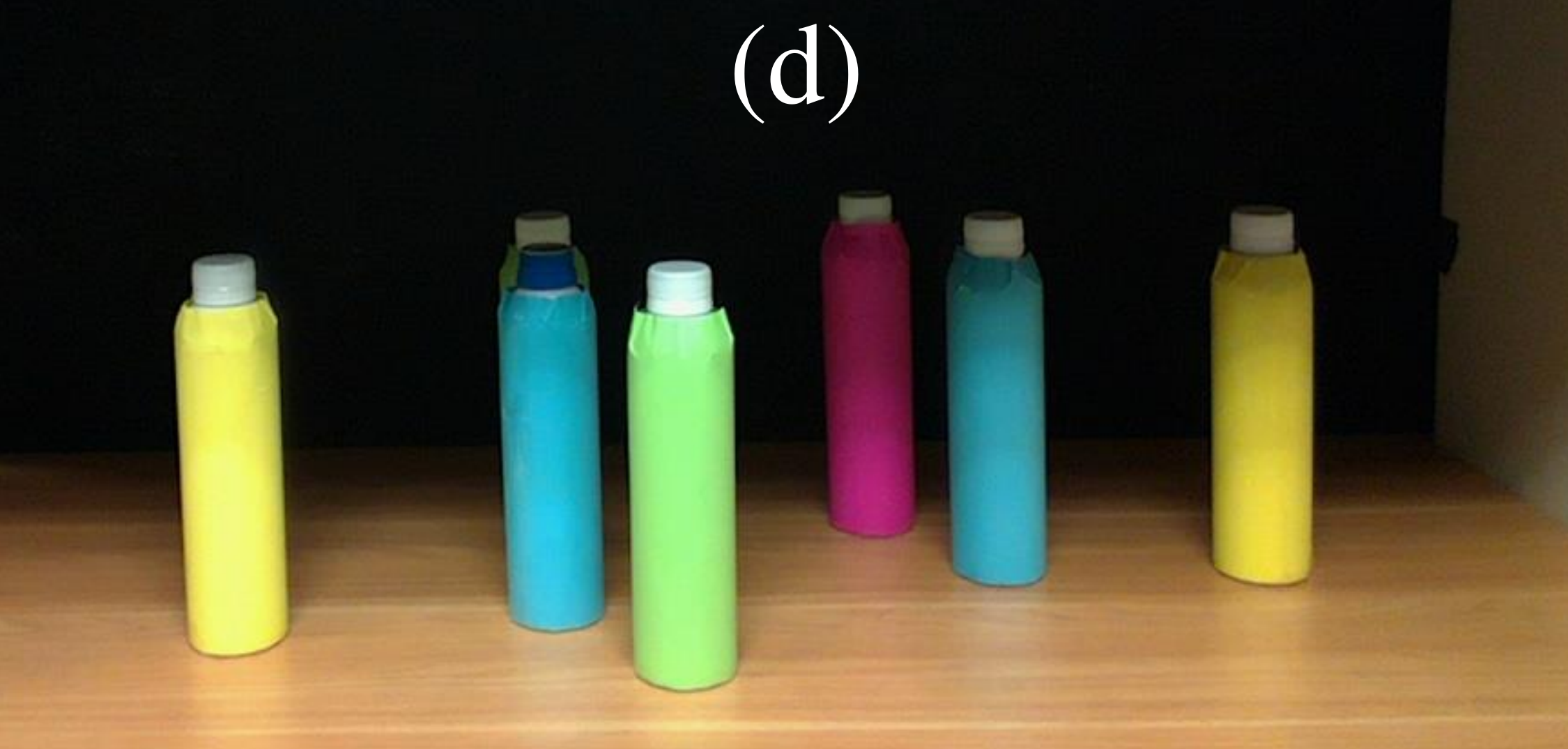}
\includegraphics[width=0.162\linewidth,trim={0 0 0 0},clip]{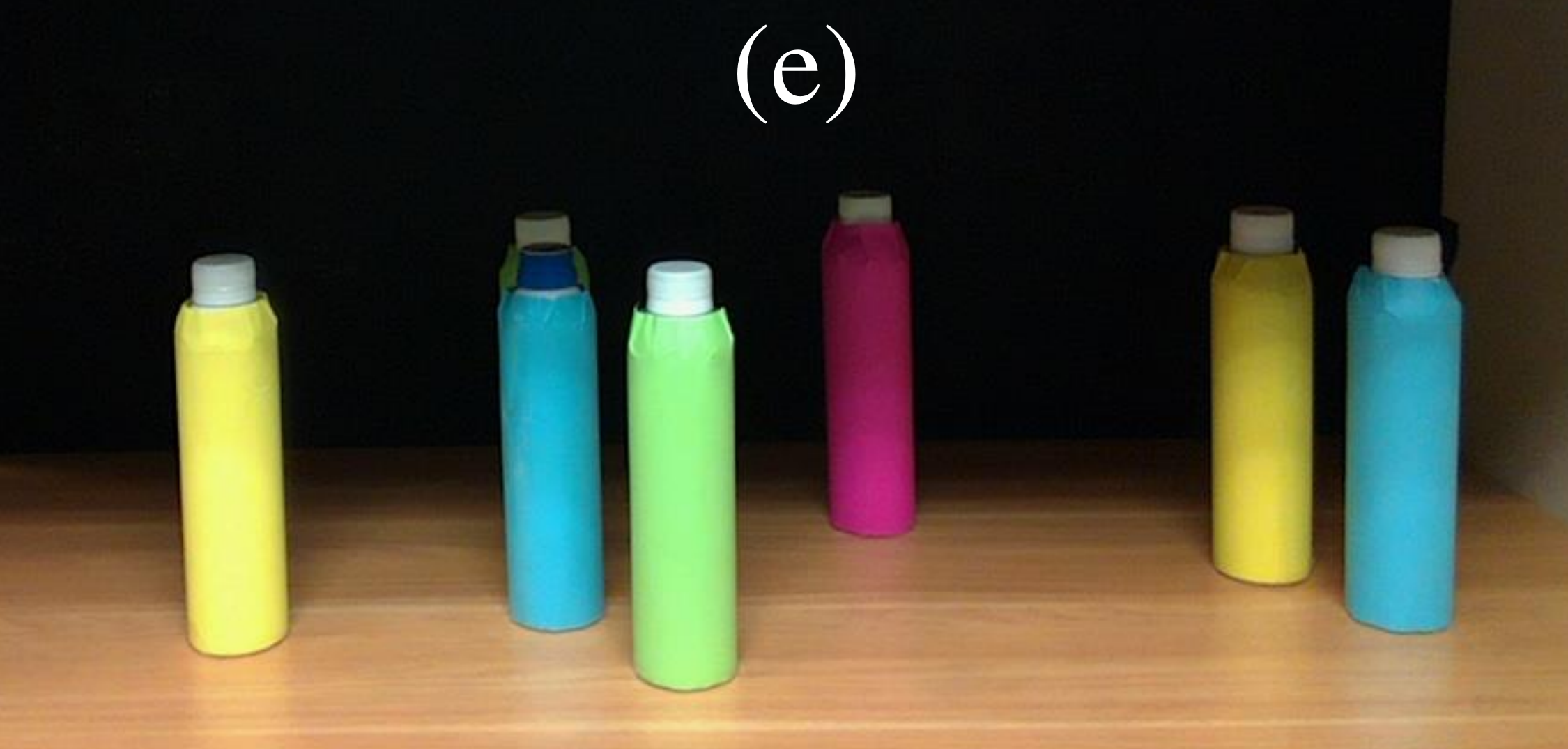}
\includegraphics[width=0.162\linewidth,trim={0 0 0 0},clip]{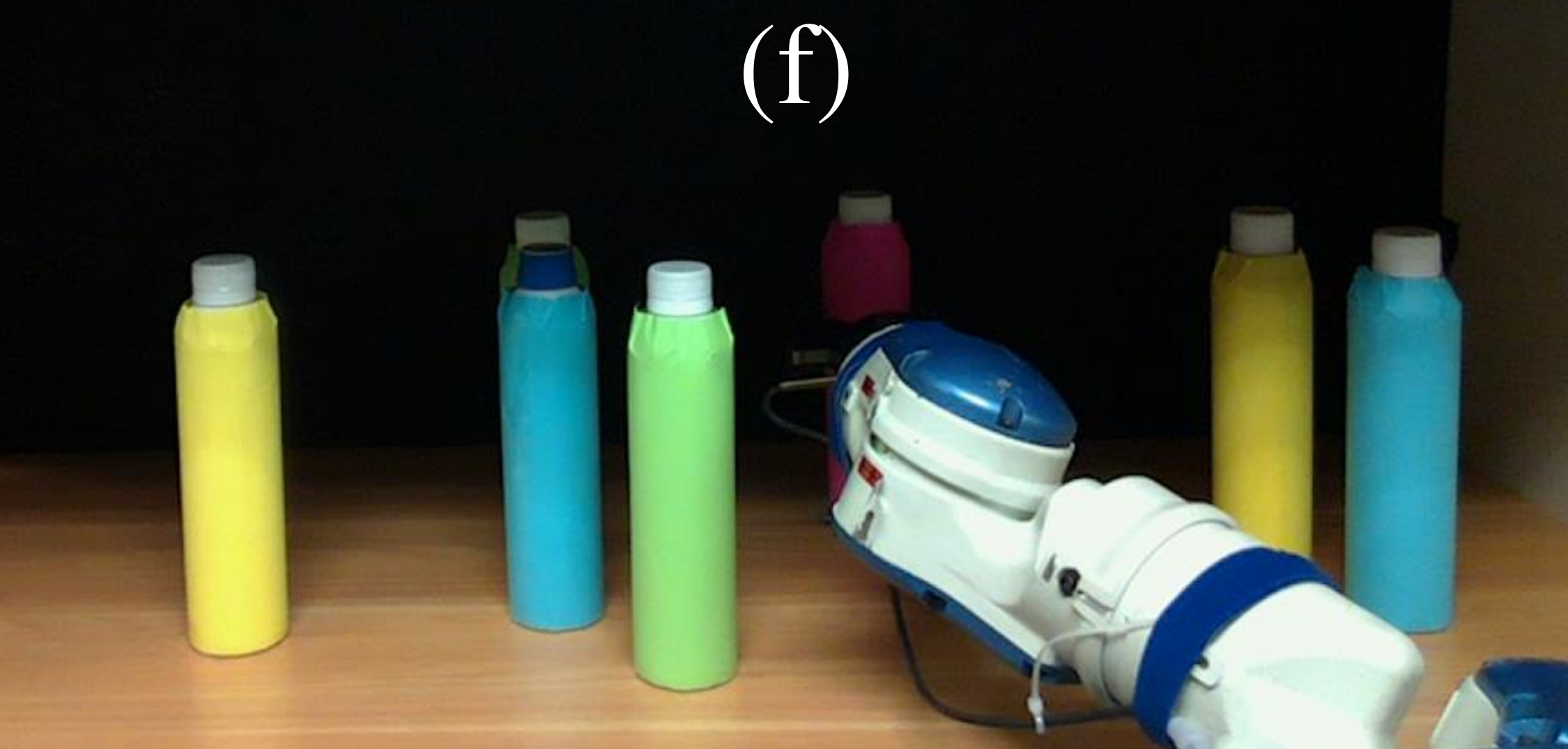}
\caption{Execution on the real robot 
: (a) Initial scene where the red bottle is hidden. (b) The robot moves the yellow bottle, which occludes the most space. (c) The robot moves the second yellow bottle, revealing the red bottle.
(d and e) The robot moves the green and the blue bottles to reach the red bottle. (f) Target is now reachable.}
\label{fig:exp-real-data}
\vspace{-0.28in}
\end{figure*}

Simulated experiments and the real demonstrations are performed with the Yaskawa Motoman sda10f, with a robotiq 85 gripper attached on the right arm.

The simulated trials are randomly generated by picking random objects, dimensions, and collision-free placements within the specified workspace. 20 scenes with each of 6, 8, 10, 12, and 14 objects were used, all of which contain objects occluded from the camera. Each of the 100 trials is a unique scene. The target object in each scene was selected to be the hidden object with the most objects above it, if any.

All tested algorithms were given 20 minutes to run before being terminated. A trial run is considered successful only if the target object was retrieved within the time limit. Discovering but failing to pick up the object was still considered a failure. 

\begin{figure}[thpb]
    \centering
    \includegraphics[width=.99\linewidth]{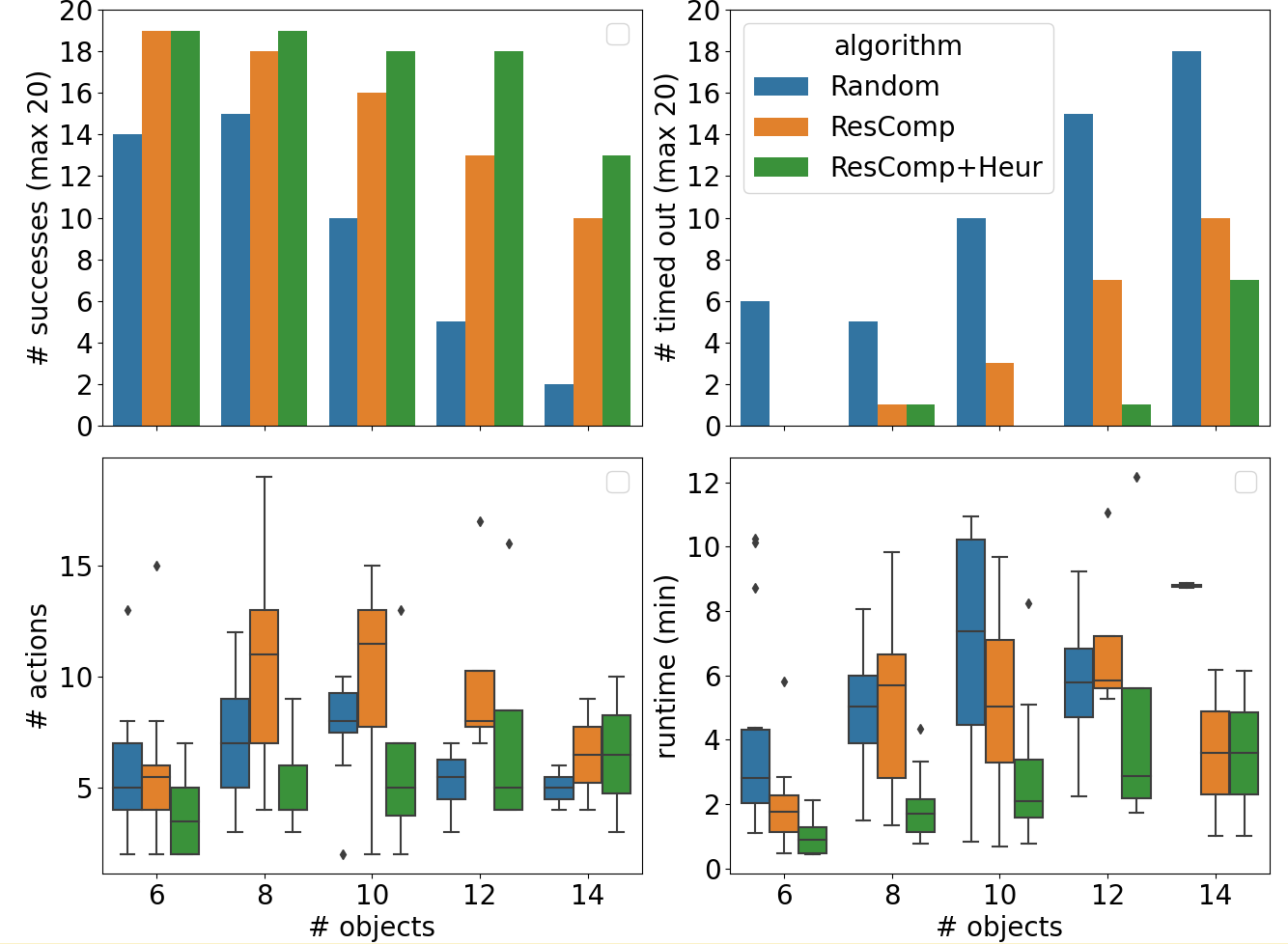}
    \caption{On top are shown the graphs of the: success rate (left) and number of timed out trials (right). On the bottom are the number of actions and the total runtime for the subset of trials in which all algorithms were successful.}
    \label{fig:exp-data}
    \vspace{-0.25in}
\end{figure}

Comparisons of the success rate, number of actions for solved trials,
total run-time for solved trials,
and number of timed out trials
are shown in \autoref{fig:exp-data} for 3 algorithms.
The algorithms compared are the baseline random action approach (blue in figure), the proposed resolution complete pipeline without the object ranking heuristic (orange) and with it (green).
For the resolution complete approaches timing-out is not the only failure mode since they could detect certain infeasible problems (\autoref{sec:completeness}: Failure Detection)

The success rate of the resolution complete approaches is higher than that of the random baseline. 
Although its not directly apparent from the plotted results in \autoref{fig:exp-data}, the resolution complete approaches (as expected) always found a solution when the random baseline found a solution; however in one such trial the resolution complete approach without heuristic exceeded the 20 min time-limit.
It is also clear that the heuristic approach has better success than the non heuristic approach even though they are both complete. Looking at the data for timed-out experiments, it becomes clear that the increased success of the heuristic approach is due to timing out less frequently. This also coincides with the data showing that the heuristic approach overwhelmingly finds solutions faster and with fewer object manipulations. In fact, while the non-heuristic {\tt RC} approach started timing out linearly with increase in the number of objects, the heuristic approach had virtually no issue until the scenes got very cluttered with 14 objects.

It is clear that for all methods, success rate starts dropping off significantly at around 14 objects.
This marks the difficulty level for the given industrial Motoman robot and the workspace.
A more compact robot with a streamlined end-effector (such as the ``bluction'' tool \cite{huang2022mechanical}) could scale to more cluttered scenes.

\subsection{Integration with Perception \& Real Robot Demonstration}

The pipeline is directly transferable to scenarios on the real robot to retrieve a target red bottle from a cluttered shelf. Due to time constraints, a simple implementation of a perception system is used which only segments and detects colored cylinders without stacking. Despite the simplifications, a scene with significant object occlusion is still demonstrated with a successful retrieval.
The proposed pipeline (with heuristic) is run online and communicates with the robot controller and the RGBD camera for execution and sensing.
The camera extrinsic matrix is estimated by a classical robot-camera calibration procedure using ArUco markers \cite{tsai1989new}.
For object recognition, the perception component is implemented via plane fitting, DBSCAN segmentation \cite{ester1996density}, and cylinder fitting using Open3D \cite{zhou2018open3d}. 
The plane fitting algorithm extracts the boundaries of the workspace, which is used to construct the collision geometries in MoveIt \cite{chitta2016moveit}.
The inliers of each segmented cylinder are used to produce the segmentation mask for the RGBD image, which is used to label occlusion correspondence for each object.
To ensure safety, additional cubic collision geometries are added to the planning scene to avoid collisions between the robot and the camera.
\footnote{Videos can be found at \url{https://sites.google.com/scarletmail.rutgers.edu/occluded-obj-retrieval}}.
Extensive experiments of the proposed pipeline were not performed on the real robot but the demonstration presented was performed a few times and the pipeline was observed to have qualitatively similar performance as in simulated experiments; however, calibration and perception issues were observed to lead to pipeline failure.

\section{DISCUSSION}

\label{sec:conclusion}

It's worth mentioning that the physical execution accounts for over 60\% of time used for the trials. This shows that there could be room for performance improvement by performing scene perception asynchronously, since a lot can still be sensed while the robot is moving. Further performance improvement can be found by parallelizing the planning of picks and placements for multiple objects as well. 

While this work applies heuristics for selecting objects based on the occlusion volume, additional information regarding effective placements can also improve practical performance. In order to solve a larger variety of problems it would be useful to adapt the placement primitive to allow placing objects on top of others when there is limited space on the workspace surface. 

Another direction is to integrate the task planner with human instructions. For instance, it would be helpful to use human language to identify the target as well as influence the search at some regions over others. Additional heuristics can also be obtained from semantic reasoning of the scene when objects of the same category tend to be placed closer \cite{li2016act}.
Since current experiments only include simple geometries, such as cylinders and rectangular prisms, future work can investigate more complex objects where state-of-the-art perception algorithms are necessary. This would also be necessary for realistic human-robot integration.







\bibliographystyle{ieeetr}
\bibliography{bib/HRI,bib/kostas,bib/stone,bib/occlusion}

\begin{thebibliography}{10}

\bibitem{wong2013manipulation}
L.~L. Wong, L.~P. Kaelbling, and T.~Lozano-P{\'e}rez, ``Manipulation-based
  active search for occluded objects,'' in {\em 2013 IEEE International
  Conference on Robotics and Automation}, pp.~2814--2819, IEEE, 2013.

\bibitem{garrett2020online}
C.~R. Garrett, C.~Paxton, T.~Lozano-P{\'e}rez, L.~P. Kaelbling, and D.~Fox,
  ``Online replanning in belief space for partially observable task and motion
  problems,'' in {\em 2020 IEEE International Conference on Robotics and
  Automation (ICRA)}, pp.~5678--5684, IEEE, 2020.

\bibitem{dogar2014object}
M.~R. Dogar, M.~C. Koval, A.~Tallavajhula, and S.~S. Srinivasa, ``Object search
  by manipulation,'' {\em Autonomous Robots}, vol.~36, no.~1, pp.~153--167,
  2014.

\bibitem{nam2021fast}
C.~Nam, S.~H. Cheong, J.~Lee, D.~H. Kim, and C.~Kim, ``Fast and resilient
  manipulation planning for object retrieval in cluttered and confined
  environments,'' {\em IEEE Transactions on Robotics}, vol.~37, no.~5,
  pp.~1539--1552, 2021.

\bibitem{ahn2021integrated}
J.~Ahn, J.~Lee, S.~H. Cheong, C.~Kim, and C.~Nam, ``An integrated approach for
  determining objects to be relocated and their goal positions inside clutter
  for object retrieval,'' in {\em 2021 IEEE International Conference on
  Robotics and Automation (ICRA)}, pp.~6408--6414, IEEE, 2021.

\bibitem{bejjani2021occlusion}
W.~Bejjani, W.~C. Agboh, M.~R. Dogar, and M.~Leonetti, ``Occlusion-aware search
  for object retrieval in clutter,'' in {\em 2021 IEEE/RSJ International
  Conference on Intelligent Robots and Systems (IROS)}, pp.~4678--4685, IEEE,
  2021.

\bibitem{huang2022mechanical}
H.~Huang, M.~Danielczuk, C.~M. Kim, L.~Fu, Z.~Tam, J.~Ichnowski, A.~Angelova,
  B.~Ichter, and K.~Goldberg, ``Mechanical search on shelves using a novel
  "bluction" tool,'' {\em IEEE International Conference on Robotics and
  Automation (ICRA)}, 2022.

\bibitem{price2019inferring}
A.~Price, L.~Jin, and D.~Berenson, ``Inferring occluded geometry improves
  performance when retrieving an object from dense clutter,'' in {\em The
  International Symposium of Robotics Research}, pp.~376--392, Springer, 2019.

\bibitem{huang2022mechanicalstack}
H.~Huang, L.~Fu, M.~Danielczuk, C.~M. Kim, Z.~Tam, J.~Ichnowski, A.~Angelova,
  B.~Ichter, and K.~Goldberg, ``Mechanical search on shelves with efficient
  stacking and destacking of objects,'' {\em arXiv preprint arXiv:2207.02347},
  2022.

\bibitem{zhu2021hierarchical}
Y.~Zhu, J.~Tremblay, S.~Birchfield, and Y.~Zhu, ``Hierarchical planning for
  long-horizon manipulation with geometric and symbolic scene graphs,'' in {\em
  2021 IEEE International Conference on Robotics and Automation (ICRA)},
  pp.~6541--6548, IEEE, 2021.

\bibitem{kumar2022graph}
K.~N. Kumar, I.~Essa, and S.~Ha, ``Graph-based cluttered scene generation and
  interactive exploration using deep reinforcement learning,'' in {\em 2022
  International Conference on Robotics and Automation (ICRA)}, pp.~7521--7527,
  IEEE, 2022.

\bibitem{liu2021ocrtoc}
Z.~Liu, W.~Liu, Y.~Qin, F.~Xiang, M.~Gou, S.~Xin, M.~A. Roa, B.~Calli, H.~Su,
  Y.~Sun, {\em et~al.}, ``Ocrtoc: A cloud-based competition and benchmark for
  robotic grasping and manipulation,'' {\em IEEE Robotics and Automation
  Letters}, vol.~7, no.~1, pp.~486--493, 2021.

\bibitem{wang_uniform_2021}
R.~Wang, K.~Gao, D.~Nakhimovich, J.~Yu, and K.~E. Bekris, ``Uniform {Object}
  {Rearrangement}: {From} {Complete} {Monotone} {Primitives} to {Efficient}
  {Non}-{Monotone} {Informed} {Search},'' in {\em 2021 {IEEE} {International}
  {Conference} on {Robotics} and {Automation} ({ICRA})}, pp.~6621--6627, May
  2021.
\newblock ISSN: 2577-087X.

\bibitem{zhao2021hierarchical}
W.~Zhao and W.~Chen, ``Hierarchical pomdp planning for object manipulation in
  clutter,'' {\em Robotics and Autonomous Systems}, vol.~139, p.~103736, 2021.

\bibitem{xiao2019online}
Y.~Xiao, S.~Katt, A.~ten Pas, S.~Chen, and C.~Amato, ``Online planning for
  target object search in clutter under partial observability,'' in {\em 2019
  International Conference on Robotics and Automation (ICRA)}, pp.~8241--8247,
  IEEE, 2019.

\bibitem{danielczuk2019mechanical}
M.~Danielczuk, A.~Kurenkov, A.~Balakrishna, M.~Matl, D.~Wang,
  R.~Mart{\'\i}n-Mart{\'\i}n, A.~Garg, S.~Savarese, and K.~Goldberg,
  ``Mechanical search: Multi-step retrieval of a target object occluded by
  clutter,'' in {\em 2019 International Conference on Robotics and Automation
  (ICRA)}, pp.~1614--1621, IEEE, 2019.

\bibitem{huang2021mechanical}
H.~Huang, M.~Dominguez-Kuhne, V.~Satish, M.~Danielczuk, K.~Sanders,
  J.~Ichnowski, A.~Lee, A.~Angelova, V.~Vanhoucke, and K.~Goldberg,
  ``Mechanical search on shelves using lateral access x-ray,'' in {\em 2021
  IEEE/RSJ International Conference on Intelligent Robots and Systems (IROS)},
  pp.~2045--2052, IEEE, 2021.

\bibitem{zeng2022robotic}
A.~Zeng, S.~Song, K.-T. Yu, E.~Donlon, F.~R. Hogan, M.~Bauza, D.~Ma, O.~Taylor,
  M.~Liu, E.~Romo, {\em et~al.}, ``Robotic pick-and-place of novel objects in
  clutter with multi-affordance grasping and cross-domain image matching,''
  {\em The International Journal of Robotics Research}, vol.~41, no.~7,
  pp.~690--705, 2022.

\bibitem{miao2022safe}
Y.~Miao, R.~Wang, and K.~Bekris, ``Safe, occlusion-aware manipulation for
  online object reconstruction in confined spaces,'' in {\em International
  Symposium on Robotics Research (ISRR)}, 2022.

\bibitem{gupta2013interactive}
M.~Gupta, T.~R{\"u}hr, M.~Beetz, and G.~S. Sukhatme, ``Interactive environment
  exploration in clutter,'' in {\em 2013 IEEE/RSJ International Conference on
  Intelligent Robots and Systems}, pp.~5265--5272, IEEE, 2013.

\bibitem{poon2019probabilistic}
J.~Poon, Y.~Cui, J.~Ooga, A.~Ogawa, and T.~Matsubara, ``Probabilistic active
  filtering for object search in clutter,'' in {\em 2019 International
  Conference on Robotics and Automation (ICRA)}, pp.~7256--7261, IEEE, 2019.

\bibitem{novkovic2020object}
T.~Novkovic, R.~Pautrat, F.~Furrer, M.~Breyer, R.~Siegwart, and J.~Nieto,
  ``Object finding in cluttered scenes using interactive perception,'' in {\em
  2020 IEEE International Conference on Robotics and Automation (ICRA)},
  pp.~8338--8344, IEEE, 2020.

\bibitem{tsai1989new}
R.~Y. Tsai, R.~K. Lenz, {\em et~al.}, ``A new technique for fully autonomous
  and efficient 3 d robotics hand/eye calibration,'' {\em IEEE Transactions on
  robotics and automation}, vol.~5, no.~3, pp.~345--358, 1989.

\bibitem{ester1996density}
M.~Ester, H.-P. Kriegel, J.~Sander, X.~Xu, {\em et~al.}, ``A density-based
  algorithm for discovering clusters in large spatial databases with noise.,''
  in {\em kdd}, vol.~96, pp.~226--231, 1996.

\bibitem{zhou2018open3d}
Q.-Y. Zhou, J.~Park, and V.~Koltun, ``Open3d: A modern library for 3d data
  processing,'' {\em arXiv preprint arXiv:1801.09847}, 2018.

\bibitem{chitta2016moveit}
S.~Chitta, ``Moveit!: an introduction,'' in {\em Robot Operating System (ROS)},
  pp.~3--27, Springer, 2016.

\bibitem{li2016act}
J.~K. Li, D.~Hsu, and W.~S. Lee, ``Act to see and see to act: Pomdp planning
  for objects search in clutter,'' in {\em 2016 IEEE/RSJ International
  Conference on Intelligent Robots and Systems (IROS)}, pp.~5701--5707, IEEE,
  2016.

\end{thebibliography}

\end{document}